\pdfoutput=1

\documentclass[11pt]{article}

\usepackage[]{acl}

\usepackage{times}
\usepackage{latexsym}

\usepackage[T1]{fontenc}

\usepackage[utf8]{inputenc}

\usepackage{microtype}

\usepackage{comment}

\usepackage{todonotes}
\usepackage{multirow}
\usepackage{setspace}
\usepackage{url}
\usepackage{booktabs}
\usepackage{adjustbox}
\newcommand*\rot{\rotatebox{90}}

\newcommand\blfootnote[1]{%
  \begingroup
  \renewcommand\thefootnote{}\footnote{#1}%
  \addtocounter{footnote}{-1}%
  \endgroup
}

\title{Twitter Topic Classification}

\author{
  Dimosthenis Antypas$^*$, Asahi Ushio$^*$, Jose Camacho-Collados \\
  Cardiff NLP, School of Computer Science and Informatics, Cardiff University, United Kingdom\\
  \texttt{\{AntypasD,UshioA,CamachoColladosJ\}@cardiff.ac.uk}\\
  \AND
  Leonardo Neves, Vítor Silva, Francesco Barbieri\\
  Snap Inc., Santa Monica, CA, United States \\
  \texttt{\{lneves,vsilvasousa,fbarbieri\}@snap.com} \\
}

\begin{document}
\maketitle


\begin{abstract}
Social media platforms host discussions about a wide variety of topics that arise everyday. Making
sense of all the content and organising it into categories is an arduous task. A common way to deal with this issue is relying on topic modeling, but topics discovered using this technique are difficult to interpret and can differ from corpus to corpus. In this paper, we present a new task based on tweet topic classification and release two associated datasets\footnote{\url{https://huggingface.co/datasets/cardiffnlp/tweet_topic_single}}\footnote{\url{https://huggingface.co/datasets/cardiffnlp/tweet_topic_multi}}. Given a wide range of topics covering the most important discussion points in social media, we provide training and testing data from recent time periods that can be used to evaluate tweet classification models. Moreover, we perform a quantitative evaluation and analysis of current general- and domain-specific language models on the task, which provide more insights on the challenges and nature of the task. 
\blfootnote{$^{*}$Equal contribution.}

\end{abstract}

\section{Introduction}

Social media platforms, e.g., Twitter, Snapchat, TikTok and Instagram, provide an environment for content creation and information sharing among people. 
On social platforms, every individual can express their views about current events or anything that they care about, influencing and guiding discussions among their friends and followers. 
Social media platforms are highly studied to understand behaviors among users, groups, organizations, or even societies \cite{Yang_Wang_Pierce_Vaish_Shi_Shah_2021}, and in particular to understand opinion of people regarding a variety of topics such as politics \cite{zhuravskaya2020political}, diversity and inclusion \cite{chakravarthi2020hopeedi}, TV shows \cite{wohn2011tweeting}, sports events \cite{lim2015social}, or finance \cite{hu2021rise}.
However, one of the biggest challenges in understanding this type of user generated content, is the noise and variety of these texts \cite{morgan2014information,baldwin2013noisy}. 
Consequently, identifying topics within social media platforms from their posts is not a trivial task.


Existing solutions can be divided into topic modeling and topic classification. For topic modeling, topics are detected in an unsupervised way with models such as Latent Dirichlet Allocation (LDA) \cite{lda} and subsequent variations \cite{steyvers2007probabilistic}. 
Similarly, solutions that use new BERT contextualized embeddings (like BERTopic \cite{bertopic}) have increased in popularity as they offer increased performance. 
However, these approaches assume that 
(i) all the topics of interest are represented in the documents included in the study, 
and (ii) the terms present in these documents are enough to characterize each topic. For these reasons, these methods are usually built as an ad-hoc analysis. 
Another limitation of these models is interpretability, as it is hard to generalize and label each cluster topic.

On the other hand, topic classification approaches the problem in a supervised manner and assigns multiple topics to each document based on a predefined set of categories. This approach overcomes the issues of interpretability and is not based on assumptions about the vocabulary distribution mentioned above. However, the downside of topic classification is that relies on curated datasets labeled by human annotators, and this can be expensive and time consuming to create.

In this paper, we introduce TweetTopic, a topic classification dataset on Twitter data.
To the best of our knowledge, this is the first large-scale topic classification dataset specifically tailored to social media, rather than standard text as news articles \cite{greene2006practical} or scientific papers \cite{lazaridou2021mind}. 
The dataset consists of a total of 11,267 tweets collected through a time period from September 2019 to August 2021. Each tweet is assigned one or more topics from a predefined set of categories curated by social platform experts. Aiming to test the robustness of our dataset through time and 
across topics, we perform several classification experiments, both single-label and multi-label, while utilizing state-of-the-art language models.\footnote{Tweet classification models associated with TweetTopic have been integrated into TweetNLP \cite{camacho2022tweetnlp}.} 

\section{Related Work}
\paragraph{Social media.}

Social media have become an important aspect of the daily life of millions of people, with 81\% of adults in the U.S. stating to have used at least one social platform 
in 2021 \cite{auxier2021social} and over 57\% of people in EU interacting through social media in 2020 \cite{eurostat}. 
In recent years, an increasing number of corporations seem to dedicate a more significant portion of their marketing funds to advertising on social platforms 
compared to other more traditional mediums \cite{eid2020internet}. At the same time, social media has become a political battleground where politicians both debate between them and try to communicate with their voters, \cite{stier2018election, llewellyn2016brexit}. 
Finally, social platforms 
have been used extensively by their users as a means for almost instantaneous news updates both for day-to-day events \cite{hermida2012social}, and human and natural disasters (e.g., the Ukrainian war or the COVID-19 pandemic) \cite{khaldarova2016fake, banda2021large}.


Therefore, a large volume of content is being generated in social media everyday.  
Its polymorphism also means that performing any targeted analysis on the data can be a challenging and time-consuming process \cite{weller2015accepting, stieglitz2018social}. Furthermore, even though there are various existing tools focused on analyzing social media data \cite{batrinca2015social}, there is no established way to efficiently identify and filter only relevant and valuable content \cite{nugroho2020survey}. 

\paragraph{Topic modeling.} Topic models are unsupervised methods to identify relevant topics given a text corpus. LDA \cite{lda} is one of the most popular algorithms for topic modeling. However, despite being successful in identifying topics in traditional media \cite{martin2015more, el2021end}, LDA often struggles when applied to short, unstructured, and constantly evolving texts, such as Twitter data \cite{zhao2011comparing}. It also typically underperforms when compared to other supervised methods \cite{arias2015ieee}. More recently, several variations of LDA have been proposed to address these challenges with social media texts, such as combining author-topic modelling with LDA \cite{rosen2004author,steinskog-etal-2017-twitter}, frameworks like Twitter-LDA \cite{zhao2011comparing} where noisy words and author information are taken into account, and SKLDA \cite{tajbakhsh2019semantic}, where semantic relations between words extracted from WordNet are taken into account.

However, LDA-based methods are often not ideal when we need to assign more than one topic to a document. Even though there are approaches to acquire multiple labels for each topic, they are usually based on hierarchical \cite{griffiths2003hierarchical} or graph \cite{li2006pachinko} architectures which, depending on the use case, make assumptions about relations of the topics that may not be present in a given corpus (i.e. parent/children topics). Furthermore, semi-supervised or supervised variations of LDA, such as PLDA \cite{ramage2011partially} and sLDA \cite{mcauliffe2007supervised}, have been been used on Twitter data \cite{resnik2015beyond, ashktorab2014tweedr}. While such methods have potential for increased performance they usually require prior labelling or information about the documents and thus remove a major advantage they have compared to supervised approaches. 

Finally, as a mainly unsupervised technique, evaluating the results of topic modeling can be a hard task. Metrics such as purity, mutual information and pairwise F-measure are used to evaluate the quality of topics/clusters created by the models \cite{nugroho2020survey}. On the other hand, qualitative analysis is usually difficult to perform due to the lack of interpretability of topics produced and the difficulty increases with the amount of topic.

In contrast to traditional LDA approaches, techniques such as BERTopic \cite{bertopic} and Top2Vec \cite{angelov2020top2vec} attempt to make use of existing knowledge from pretrained language models by extracting embedding representations of tweets and using them to perform topic clustering. Both BERTtopic and Top2Vec tend to be easier to use than LDA, without the need for extensive hyper-parameter tuning, and often result in increased performance \cite{egger2022topic}. However, they do have disadvantages, namely: not performing well on small datasets \cite{abuzayed2021bert}, generating a lot of outlier topics \cite{silveira2021topic}, and requiring existing knowledge. Finally, these approaches suffer similar drawbacks to LDA regarding evaluation and interpretability.

\paragraph{Topic classification.}
Given a text as an input, topic classification is the task of associating it with a specific topic (or topics) from a pre-defined set of categories. In what concerns social media, previous work has focused on predicting hashtags as classes \cite{tweet2vec}. However, the dynamic nature of the events discussed in those platforms makes any dataset focused on hashtags quickly become sparse and outdated. Any new model needs to be trained from scratch since the category set will be different based on the relevance of hashtags. 
Nevertheless, by focusing on higher-level topics like \emph{Sports} or \emph{Arts \& Culture}, widespread and recurrent in social platforms, the data can be leveraged for more extended periods, and any model trained on it can be easily updated with more data as the label set is fixed. It also improves interpretability since there is a clear semantic meaning to the proposed categories, while hashtags might be ambiguous or require additional interpretation.

In terms of previously released data, existing datasets mainly focus on the news articles domain, e.g., BBC News \cite{greene2006practical}, Reuter \cite{lewis2004rcv1}, 20 Newsgroups \cite{lang1995newsweeder}, and WMT News Crawl \cite{lazaridou2021mind} with few exceptions like scientific (arXiv) \cite{lazaridou2021mind} and medical (Ohsumed) \cite{hersh1994ohsumed} domains. Therefore, these datasets offer different sets of challenges with respect to social media.


\section{Tweet Topic Classification} 
\label{data}

This section presents the pipeline to construct TweetTopic, our topic classification dataset based on Twitter data. 
This pipeline is divided into three steps: (i) tweet collection, (ii) data filtering, and (iii) topic annotation. 
These steps are explained in more detail in the next subsections.

\subsection{Tweet collection}
Our goal is to collect a set of tweets with a high coverage of diverse topics over time. We fetched the tweets given specific keywords and time periods using the Twitter API. 
Since the tweets returned by the API are in reverse chronological order, we decided to split the queries into small time windows to make sure that the tweets are distributed over time. In our case, we queried 50 tweets every two hours from September 2019 to October 2021.
As the keywords used to create queries, we collected lists of trending topics from Snapchat\footnote{Available at \url{https://trends.snapchat.com/}. We were not able to access Twitter trends since they are not publicly available through APIs.} in each week during the period
(e.g. \emph{pink super moon}, \emph{social distancing}, and \emph{NBA}).
This step allowed us to collect tweets with a similar distribution to topics in the real world over time. 
For this step we also added conditions to exclude retweets, replies, quotes, and tweets with media, as well as specifying the language as English only.
In the end, we collected a total of 1,264,037 raw tweets from the API.

\begin{figure}[t]
    \centering
    \includegraphics[width=7.5cm]{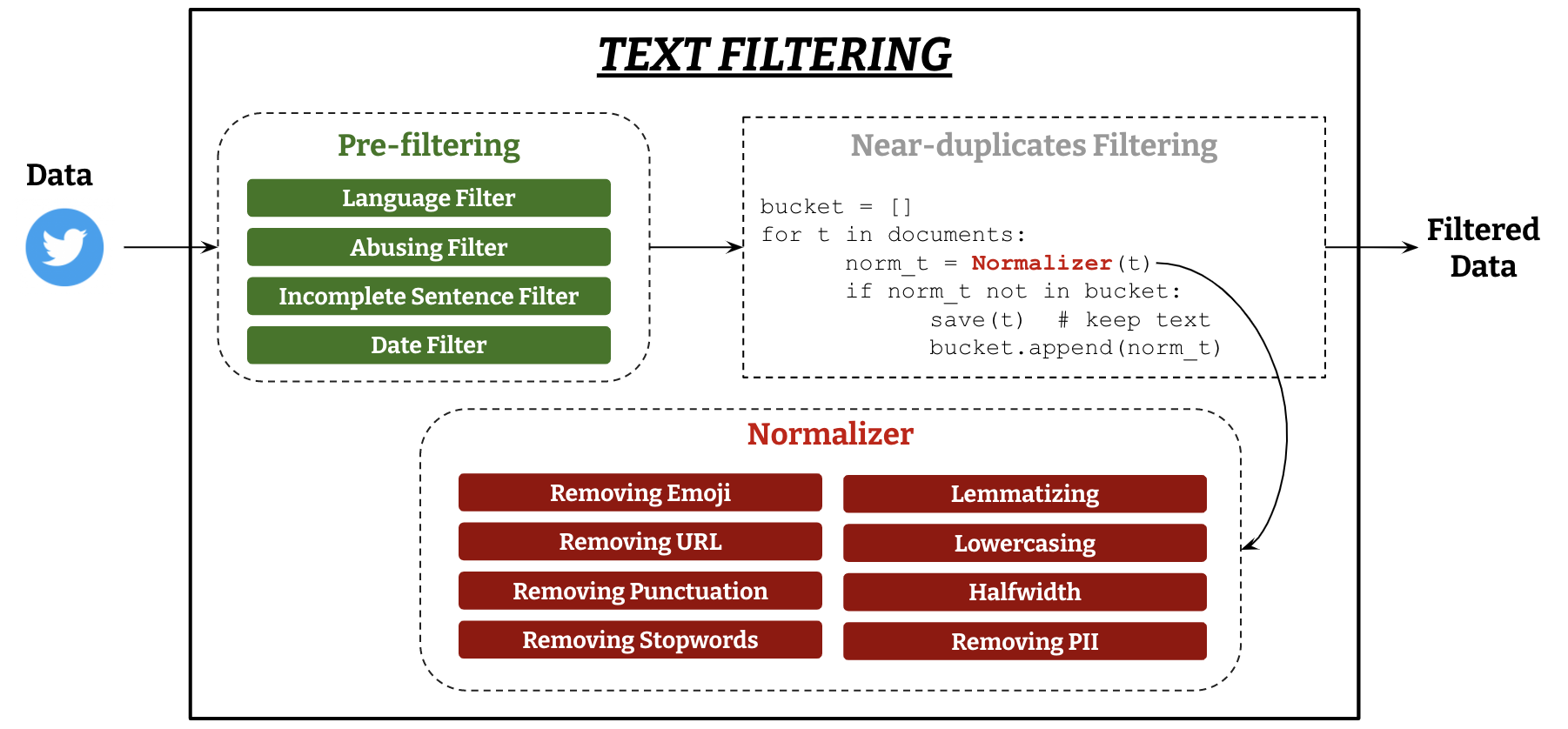}
    \caption{Text filtering pipeline to reduce noise from the tweets and avoid near duplicates.}
    \label{fig:filtering}
\end{figure}

\subsection{Data Filtering}

\paragraph{Tweet filtering.} Since the raw tweets may contain irrelevant content, we applied several text filtering techniques to get a cleaner tweets corpus.
Our text filtering pipeline consists of two steps as described in Figure~\ref{fig:filtering}: \textit{pre-filtering} and \textit{near-deduplication}.
This filtering fulfilled different goals such as removing abusive content, improving quality and avoiding near-duplicates.
In the \textit{pre-filtering}, we first removed non-English tweets by using a fastText based language identifier\footnote{\url{https://fasttext.cc/blog/2017/10/02/blog-post.html}}  \cite{bojanowski2016enriching}. 
Then, we removed tweets that 
contained incomplete sentences (e.g., too short or end in the middle of the sentence) or abusing words by using rule-based heuristics. 
Then, we applied a \textit{near-duplication} filter to drop duplicated tweets. 
In particular, we first normalized each tweet, and kept unique tweets only in terms of their normalized form.
The normalizer first converted full-width to half-width and removed substrings from the tweet such as emoji, web URLs, punctuation, stopwords, and personally identifiable information (PII).\footnote{We detected PII with scrubadub 
and other components are all based on NLTK.}  
Then, we lemmatized and lowercased each word in the tweets and removed identical tweets after normalization.

\paragraph{Trend filtering.}
Given our budget and in order to further reduce the number of tweets to annotate while ensuring diversity, we grouped the tweets by the trending topics used to query the raw tweets in each week, and selected the top 15 most common trends within the week.\footnote{More details about this process can be found in the Appendix.} 
We applied the trending topic filtering for every week which resulted in our final dataset, consisting of 28,573 tweets in total. 
Note that the trends are different every week, so the tweets are diverse across weeks regarding the trends.\footnote{In the Appendix we provide a detailed breakdown of the distribution of trends in each week. There, we can confirm that the top trend does not go beyond 20\% in most cases, which ensures a diverse set of trends.
} 

\subsection{Annotation}

To attain topic annotations over the tweets, we conducted a manual annotation on Amazon Mechanical Turk.
We randomly sampled 11,374 tweets from the cleaned tweets and each tweet was annotated by five annotators, collecting 56,870 annotations in total. 
We manually constructed a topic taxonomy that contained 23 initial topics across diverse genres, asked workers to annotate the relevant (possibly multiple) topics to the tweet.\footnote{The actual instructions shown to workers are included in the Appendix.} 
The initial list of 23 topics was shared with us by a research team of Snapchat. 
This list was selected and curated by a team of social media experts from the company over time to ensure a tailored coverage of social media content. 

We ensured several quality control mechanisms within the test, including a qualification test. Each tweet was annotated by five turkers and the final budget for the total estimated annotation cost was \$4,000. 
Each single assignment contained 50 tweets to be annotated where each annotation is completed with an interface that we include in the Appendix. 
As quality control, each assignment contained three qualification tweets and only those who annotated them correctly were accepted.
A small number of raters (10) and their respective tweets were also discarded as they displayed unusual behavior selecting on average more than 5 labels for each tweet where the global average was 1.6 labels per tweet.
Also, workers were not allowed to work on the assignment more than once.

\paragraph{Post-aggregation.} 
We followed \newcite{mohammad2018semeval} by assigning a label to a tweet provided that the label was suggested by at least two annotators. 
We opted out of a majority rule as this way our dataset can be used to develop more robust systems that can handle real-world data, which are rarely straightforward and instead can often contain complex linguistic phenomena \cite{mohammad2018semeval}. Tweets where none of the classes received at least two votes were discarded.
The number of tweets in each process is summarized in Table~\ref{tab:num-tweets}.

\begin{table}[t]
\centering
\scalebox{0.75}{
\begin{tabular}{@{}c@{\hspace{5pt}}c@{\hspace{5pt}}c@{\hspace{5pt}}c@{\hspace{5pt}}c@{}}
\toprule
      Raw       & Pre-filter & De-duplication & Trend-filter & Annotated  \\\midrule
 1,264,037 & 596,028    & 202,604        & 28,573 & 11,267 \\\midrule
\bottomrule
\end{tabular}
}
\caption{Number of total tweets after each step.}
\label{tab:num-tweets}
\end{table}

\paragraph{Inter-annotator agreement.}
Several metrics can be used to evaluate the quality of an annotation task \cite{artstein2008inter} and it is often difficult to select the most appropriate one. In our experiment, 
we utilized Krippendorff's alpha \cite{krippendorff2011computing} with MASI distance \cite{passonneau2006measuring}, which is a common combination when dealing with multi-rater and multi-label tasks \cite{artstein2008inter}. For our task the alpha statistic results in 0.35. As a comparison reference, a completely random annotation would produce a 0 alpha statistic. 
When considering the percent agreement of each pair of annotators we acquire a value of 0.87 in contrast to 0.62 for random annotation. 
These inter-annotator agreement results appear to be inline or slightly better then previous similar multi-label annotation tasks \cite{mohammad2018semeval}.

\subsection{Settings and temporal split}
\label{settings}

In order to investigate potential temporal differences in the corpus we split the datasets into two periods: (1) from September 2019 to August 2020 (referred to as training data) and (2) from September 2020 to August 2021 (test data). The motivation behind this temporal split is to make the task more realistic and evaluate the generalizability performance of the classifiers on future data.

We established two classification settings: (1) multi-label and (2) single-label. Sample instances from both settings are displayed in Table \ref{table:tweet_examples}.\footnote{For readability, tweet examples have been slightly modified within the paper, removing links and usernames which are anonymized in the dataset.} With this distinction, we aim to provide flexibility to users, and increase the usability of the dataset for settings and analyses, where a more fine-grained classification of tweets is not required (i.e. single-label).

\begin{table}[t]
\adjustbox{max width=\columnwidth}{%
\begin{tabular}{l|l}
\multicolumn{1}{c}{\textbf{Tweet}} &
  \multicolumn{1}{c}{\textbf{Topics}} \\
\begin{tabular}[c]{@{}l@{}}Apple Removed More Than 30,000 \\ Apps From The Chinese App Store \end{tabular} &
  \begin{tabular}[c]{@{}l@{}}- bus \& ent\\ - news \& soc\\ - sci \& tech\end{tabular} \\ \hline
\begin{tabular}[c]{@{}l@{}}\#copreps Football: \\ End of the line for FLHS season\end{tabular} &
  sports \& games
\end{tabular}
}
\caption{Sample tweets for each setting studied (top: multi-label; botttom: single-label).} 
\label{table:tweet_examples}
\end{table}

\paragraph{Multi-label.} 
By applying a final post-aggregation step to exclude categories that may not be relevant for social media, we removed those categories with fewer than 50 labels overall, leaving a final set of 19 topics. 

\paragraph{Single-label.} In an effort to keep the classes relatively balanced, we firstly excluded tweets that were labeled with the most dominant of the classes, i.e., news \& social concern (32.82\% of total tweets), which is highly cross-category. Following this, the remaining ten most prominent classes were considered. Finally, based on logical assumptions regarding the similarity of the classes and also the overlap between them, several labels were grouped together. 
More specifically: \textit{gaming} and \textit{sports} (35\% overlap) were grouped as \textit{sports \& gaming}; \textit{music}, \textit{celebrity \& pop culture}, and \textit{film tv \& video} (44\% and 31\% overlap)  became \textit{pop culture}; \textit{diaries \& daily life} and \textit{family} (54\% overlap) were grouped together as \textit{daily life}. These three new classes along with the original \textit{arts \& culture}, \textit{business \& entrepreneurs}, and \textit{science \& technology} composed the final set of topics. Finally, in this setting, tweets containing more than one of these six labels were dropped.

\subsection{Statistics}

\begin{figure}[t]
    \centering
    \includegraphics[width=7.5cm]{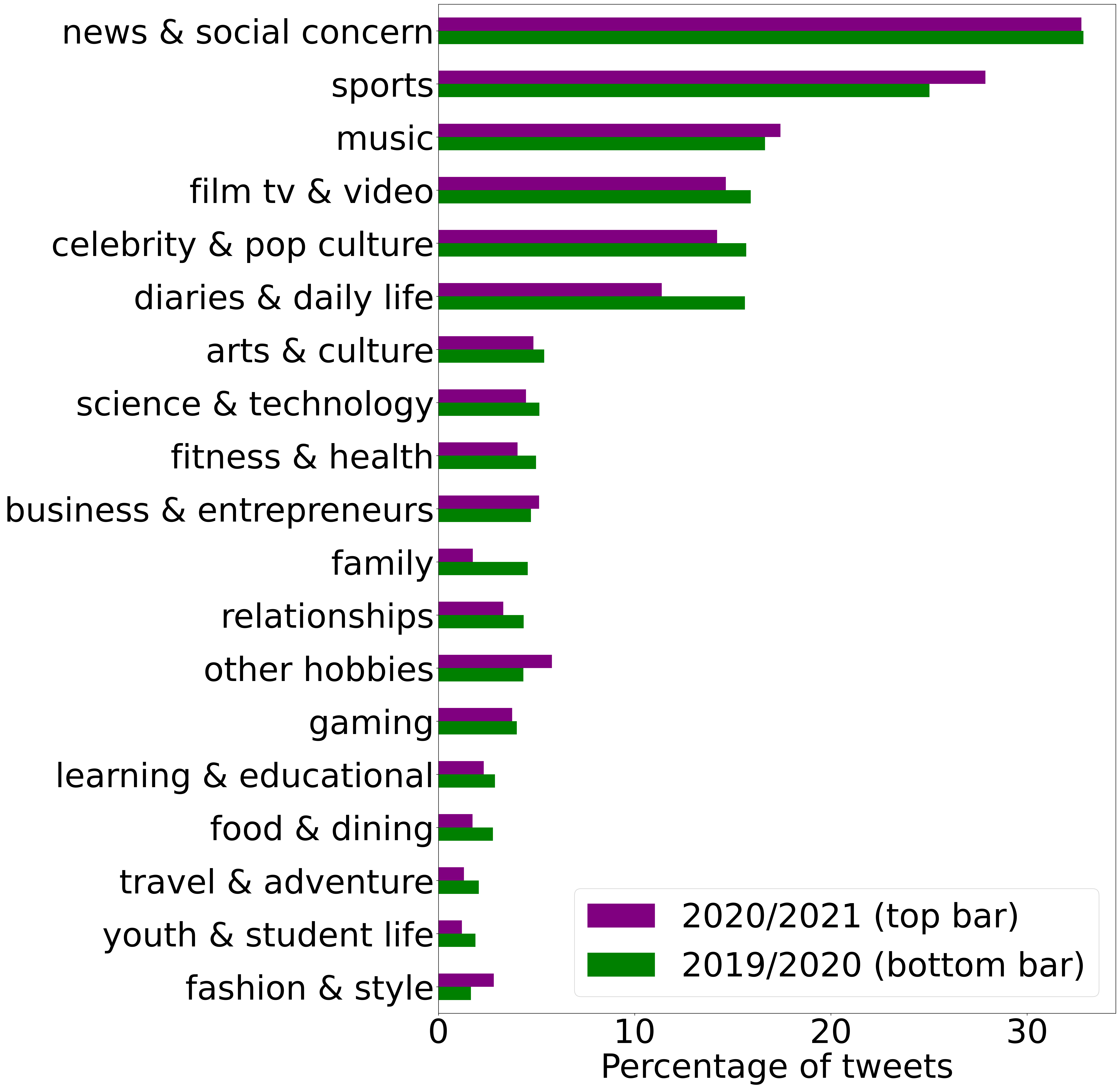}
    \caption{Percentage of tweets that were annotated with a given topic (multi-label setting) for each time period.}
    \label{fig:class_dist_multilabel}
\end{figure}

\begin{figure}[t]
    \centering
    \includegraphics[width=7.5cm]{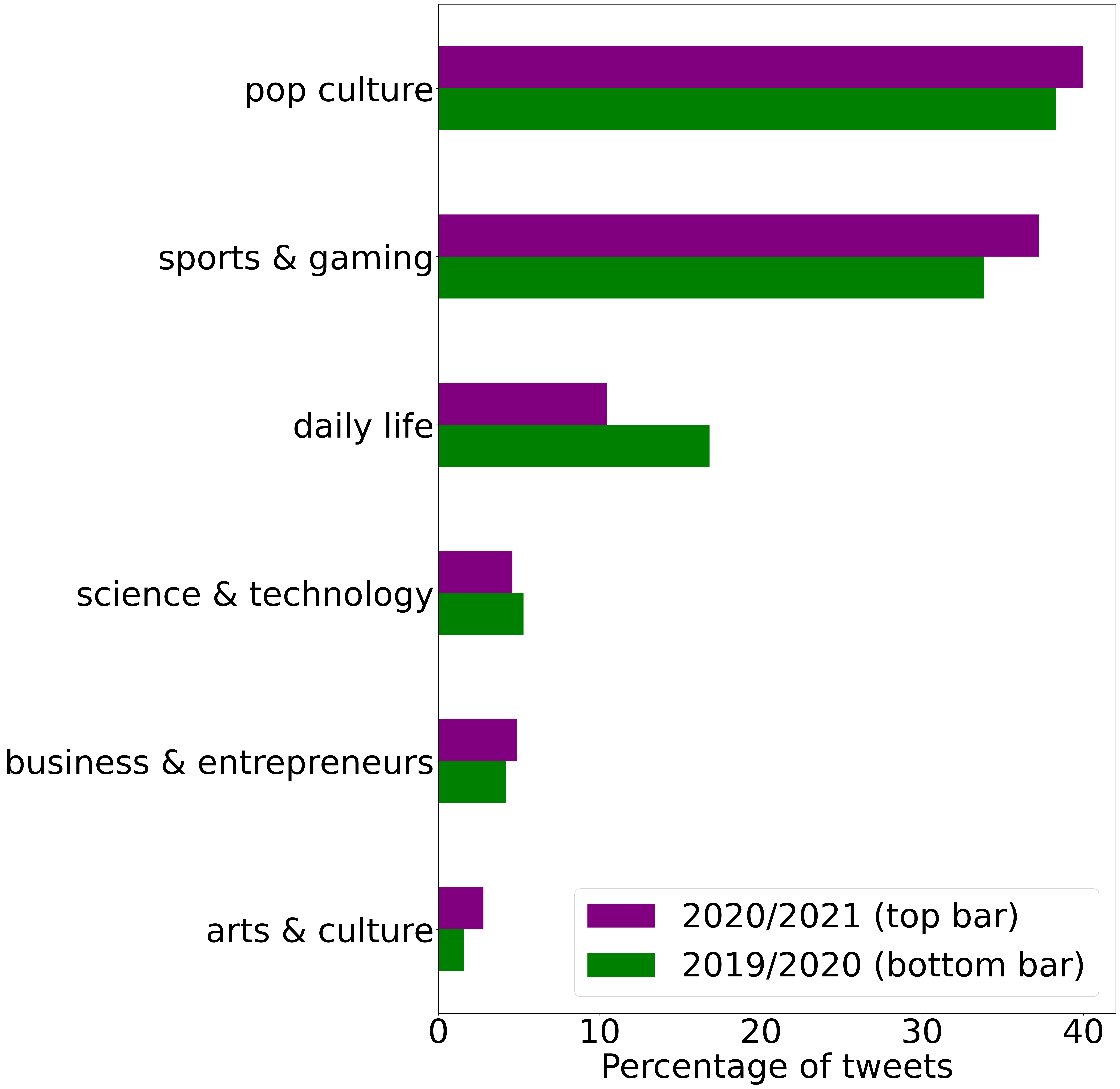}
    \caption{Percentage of tweets that were annotated with a given topic (single-label setting) for each time period.}
    \label{fig:class_dist_singleclass}
\end{figure}

The final set of annotated tweets is 11,267 and 6,997 for the multi-label and single-label settings, respectively. Figures \ref{fig:class_dist_multilabel} and \ref{fig:class_dist_singleclass} display the percentage of tweets that were classified in each topic, for each time period studied, after the aggregation of annotations for multi-label and single-label settings, respectively.\footnote{For the multi-label setting the percentages sum up to more than 100\% due to the nature of the annotation.} The imbalanced nature that can be observed, e.g., \textit{sports} consisting of 26\% of the 2019/20 multi-label dataset while \textit{travel \& adventure} only 2\%, is explained due to the way tweets were collected, where we aimed to mimic the distribution of real-world data on Twitter.


\paragraph{Number of labels.} When considering the multi-label setting, 50\% of the tweets are classified with only one label while only 2.7\% are given four or more labels, with the maximum amount being six. However, the dataset is diverse enough 
with 35\% and 12\% of the tweets having two and three labels respectively. This coder behavior (i.e. preferring to select only one class) can be observed on similar multi-label annotation tasks \cite{veronis1998study,poesio2005reliability}.

\begin{table*}[ht]
\adjustbox{max width=\textwidth}{%
\begin{tabular}{|l|c|c|c|c|c|c|c||c|}
\hline
\textbf{Class} & \textbf{length} & \textbf{punc} & \textbf{upp/low} & \textbf{\#} & \textbf{@} & \textbf{emojis} & \textbf{mtld} & \textbf{count}\\ \hline
arts \& culture           & 166.9 \textpm 67.5 & 6.5 \textpm 3.4 & 0.2 \textpm 0.6 & 0.8 \textpm 1.4 & 0.4 \textpm 0.5 & 0.1 \textpm 0.3 & 140.9 & 577 \\ \hline
business \& entrepreneurs & 186.3 \textpm 65.5 & 6.4 \textpm 3.1 & 0.1 \textpm 0.2 & 0.6 \textpm 1.1 & 0.5 \textpm 0.5 & 0.0 \textpm 0.2 & 159.0 & 554 \\ \hline
celebrity \& pop culture & 155.5 \textpm 67.8 & 7.4 \textpm 3.7 & 0.2 \textpm 0.9 & 0.6 \textpm 1.0 & 0.8 \textpm 0.7 & 0.1 \textpm 0.4 & 145.8 & 1685 \\ \hline
diaries \& daily life    & 168.3 \textpm 68.4 & 5.4 \textpm 3.3 & 0.1 \textpm 0.7 & 0.4 \textpm 0.9 & 0.4 \textpm 0.5 & 0.1 \textpm 0.5 & 132.5 & 1525 \\ \hline
family                      & 165.1 \textpm 68.5 & 5.2 \textpm 3.2 & 0.2 \textpm 1.4 & 0.5 \textpm 1.0 & 0.4 \textpm 0.5 & 0.2 \textpm 0.5 & 112.7 & 358 \\ \hline
fashion \& style          & 147.9 \textpm 55.4 & 7.8 \textpm 3.1 & 0.2 \textpm 0.5 & 1.0 \textpm 1.5 & 0.6 \textpm 0.5 & 0.1 \textpm 0.3 & 98.8  & 251 \\ \hline
film tv \& video         & 157.7 \textpm 66.3 & 7.5 \textpm 3.7 & 0.2 \textpm 0.8 & 0.6 \textpm 1.1 & 0.7 \textpm 0.6 & 0.1 \textpm 0.4 & 145.1 & 1723\\ \hline
fitness \& health         & 195.4 \textpm 67.1 & 6.3 \textpm 2.8 & 0.1 \textpm 0.1 & 0.5 \textpm 0.9 & 0.6 \textpm 0.5 & 0.1 \textpm 0.3 & 168.5 & 508 \\ \hline
food \& dining            & 165.2 \textpm 64.5 & 6.1 \textpm 3.1 & 0.1 \textpm 0.2 & 0.5 \textpm 1.0 & 0.4 \textpm 0.5 & 0.1 \textpm 0.4 & 154.7 & 255 \\ \hline
gaming                      & 159.6 \textpm 68.9 & 6.5 \textpm 3.9 & 0.1 \textpm 0.2 & 0.5 \textpm 1.0 & 0.5 \textpm 0.6 & 0.0 \textpm 0.2 & 128.4 & 437 \\ \hline
learning \& educational   & 191.8 \textpm 65.8 & 5.9 \textpm 2.9 & 0.1 \textpm 0.1 & 0.6 \textpm 1.0 & 0.5 \textpm 0.6 & 0.0 \textpm 0.2 & 156.7 & 293 \\ \hline
music                       & 143.5 \textpm 64.0 & 8.4 \textpm 4.4 & 0.3 \textpm 1.1 & 0.7 \textpm 1.1 & 0.8 \textpm 0.7 & 0.1 \textpm 0.5 & 119.8 & 1919\\ \hline
news \& social concern   & 183.1 \textpm 70.5 & 6.6 \textpm 3.0 & 0.2 \textpm 1.3 & 0.4 \textpm 0.8 & 0.6 \textpm 0.6 & 0.0 \textpm 0.2 & 165.1 & 3698 \\ \hline
other hobbies              & 160.9 \textpm 69.2 & 6.3 \textpm 3.4 & 0.2 \textpm 0.7 & 0.6 \textpm 1.0 & 0.4 \textpm 0.6 & 0.1 \textpm 0.4 & 143.6 & 568 \\ \hline
relationships               & 162.4 \textpm 70.6 & 5.3 \textpm 3.5 & 0.2 \textpm 1.6 & 0.4 \textpm 0.9 & 0.5 \textpm 0.6 & 0.2 \textpm 0.9 & 111.9 & 432 \\ \hline
science \& technology     & 177.9 \textpm 69.4 & 6.7 \textpm 2.8 & 0.1 \textpm 0.5 & 0.5 \textpm 1.0 & 0.6 \textpm 0.5 & 0.0 \textpm 0.1 & 164.2 & 542 \\ \hline
sports                      & 162.8 \textpm 65.9 & 6.4 \textpm 3.2 & 0.2 \textpm 1.4 & 0.5 \textpm 0.8 & 0.7 \textpm 0.6 & 0.1 \textpm 0.3 & 152.8 & 2977 \\ \hline
travel \& adventure       & 175.2 \textpm 72.3 & 6.2 \textpm 3.1 & 0.2 \textpm 1.8 & 0.5 \textpm 1.0 & 0.5 \textpm 0.5 & 0.1 \textpm 0.2 & 173.1 & 190 \\ \hline
youth \& student life    & 202.0 \textpm 62.4 & 5.9 \textpm 3.2 & 0.1 \textpm 0.1 & 0.5 \textpm 0.9 & 0.5 \textpm 0.6 & 0.1 \textpm 0.2 & 155.6 & 174 \\ \hline
\end{tabular}}
\caption{General lexical statistics for each class. The averages of the length of tweet, punctuation count, upper/lower case ratio (upl/low), hashtags count, mentions count, emojis count are reported along with their standard deviation. Frequency metrics are normalized based on the text length. 
The last two columns correspond to the lexical diversity (mtld) and total number of tweets.}
\label{tab:corpus_statistics}
\end{table*}

\paragraph{Class distribution across time periods.} We note that the distribution of classes between the two time periods studied remains largely similar in both settings with the largest difference being in the \textit{music} and \textit{news \& social concern} classes being 3.5\% more populous in 2019/20. This observation suggests that our curated topics are broad enough to be relatively robust to temporal trends.

\paragraph{Topic features.} In order to get a better understanding of the data, and to investigate potential significant characteristics, we extract various statistics from the tweets in the multi-label dataset. Table \ref{tab:corpus_statistics} displays the average values of tweet length, number of punctuation symbols, upper to lower case ratio, number of hashtags, number of mentions and number of emojis, along with their standard deviations for each topic. In order to have a fair comparison, all the metrics are normalized based on the tweet length ($(metric/length) * 100$). 
The Measure of Textual Lexical Diversity (MTLD) \cite{mccarthy2010mtld} is also reported as an indication on the vocabulary richness of each class, as well as the number of tweets for each class. 
The topics \textit{celebrity \& pop culture} and \textit{music} have the highest occurrences of mentions "@" (0.8). This is intuitively due to the fact that a large number of tweets belonging to these classes will mention recognizable users such as artists or athletes. Similarly, tweets belonging to the \textit{fashion \& style} topic tend to include more hashtags (\#) on average (1 hashtag per tweet), which can be attributed to the nature of hashtags in Twitter, usually employed to indicate popular and trending topics. 
Finally, topics that can be considered more accessible to the general public such as \textit{fashion \& style}, \textit{family}, and \textit{relationships} achieve a relatively low lexical diversity score (98.8, 112.7, 111.9) while more specialized or advanced topics such as \textit{travel \& adventure}, \textit{business \& entrepreneurs} and \textit{fitness \& health} display higher lexical diversity (173.1, 159.0, 168.5).

\section{Evaluation}
\label{evaluation}

In this section, we present our experimental results.

\subsection{Experimental setting}

\noindent \textbf{Datasets.} We perform experiments in our tweet classification annotated datasets. In particular, our experiments are based on two settings, single-label and multi-label (see Section \ref{settings} for details). 




\noindent \textbf{Comparison systems.} To evaluate our dataset, we first use simple baselines: Majority (most frequent class in training) and Random (uniform probability for each class). As comparison systems, we train a traditional bag of words with SVM
and a fastText classifier \cite{bojanowski2016enriching} that utilizes pretrained embeddings \cite{mikolov2018advances}. Furthermore, BERT base and large \cite{DBLP:journals/corr/abs-1810-04805} and both base and large versions of RoBERTa \cite{DBLP:journals/corr/abs-1907-11692} are used as comparison systems. As classifiers specialized on social media, i.e. trained on Twitter data, BERTweet \cite{bertweet}, TimeLM-19, and TimeLM-21 \cite{loureiro2022timelms}, all based on a RoBERTa architecture, are also utilized. BERTweet is trained on a corpus of 845M tweets mainly from 01/2012 to 08/2019, while also including 5M COVID-19 related tweets from 01/2020 to 03/2020. On the other hand, TimeLM-19 is trained on 95M tweets gathered between 2018 and 2019. For completeness, we also report results of TimeLM-21, trained on 125M tweets from 2018 to 2021, but excluded it from our main analysis given the time overlap with the test set (reminder that one of the motivations of this task is to be able to process tweets in real time). TimeLMs models use the RoBERTa-base model as initial checkpoint, while BERTweet is trained from scratch. 
The implementations provided by Hugging Face \cite{wolf2019huggingface} are used to train and test all language models.\footnote{More details about the exact hyperparameters are included in the Appendix.}

\noindent \textbf{Evaluation metrics.} For both settings macro average Precision, Recall and F1, as well as Accuracy, are used to evaluate the models tested. As an alternative  metric for the multi-label setting, Jaccard Index (JI) is also utilized, as it can offer useful insights about the models performances \cite{pereira2018correlation,tsoumakas2009mining}. More specifically, the index is calculated for each tweet individually and the final metric is computed as the average over all entries. 

\subsection{Results}
Table \ref{table:metrics} displays the results of all comparison system on both settings. While only a number of models were tested, the results suggest that domain-specific knowledge appears to be more important than the size of the model, with Twitter base models outperforming large generic language models. Given the larger number of labels and more challenging setting, multi-label classification appears to be most challenging setting with the best model TimeLM-21, barely achieving 58.8\% F1 and 67.6\% Jaccard scores, in comparison to 70.1\% F1 and 86.8\% Accuracy in the single-label setting. However, it is important to note that TimeLM-21 has the unfair advantage of being trained with a more recent corpus and more specifically a corpus from the same time period as the test set. Taking this into consideration, the next best performing model is TimeLM-19 with 57.2\% and 70\% F1 scores, for the multi-label and single-label settings respectively. Even though the differences in the average F1 scores between the two models is relatively small, 1.6\% and 0.1\% for multi/single settings, when taking into account their performance in each individually topic, we can identify topics where TimeLM-21 clearly outperforms TimeLM-19 (see Section \ref{sec:temporal} for more details).

\begin{table}[t!]
\centering
\adjustbox{max width=\columnwidth}{%
\setlength{\tabcolsep}{1.7pt}
\begin{tabular}{c|l|ccccc|cccc}
\multirow{2}{*}{\textbf{\begin{tabular}[c]{@{}c@{}}\end{tabular}}} &
  \multicolumn{1}{c}{\multirow{2}{*}{\textbf{Model}}} &
  \multicolumn{5}{c}{\textbf{Multi-label}} &
  \multicolumn{4}{c}{\textbf{Single-label}} \\
 &
  \multicolumn{1}{c}{} &
  \textbf{Pr} &
  \textbf{Rec} &
  \textbf{F1} &
  \textbf{Acc} &
  \textbf{JI} &
  \textbf{Pr} &
  \textbf{Rec} &
  \textbf{F1} &
  \textbf{Acc} \\  \hline
\multirow{4}{*}{\rot{\textbf{Baselines}}} &
  Random &
  8.4 &
  48.3 &
  12.6 &
  0 &
  7.9 &
  15 &
  14.2 &
  11.9 &
  15.5 \\
 &
  Majority &
  \multicolumn{1}{c}{1.5} &
  \multicolumn{1}{c}{5.3} &
  \multicolumn{1}{c}{2.3} &
  18.0 &
  22.6 &
  6.7 &
  \multicolumn{1}{c}{16.7} &
  \multicolumn{1}{c}{9.5} &
  40 \\
 &
  SVM &
  69.4 &
  23.7 &
  30.5 &
  37.1 &
  51.8 &
  73.6 &
  47.4 &
  50.2 &
  75.8 \\
 &
  fastText &
  \multicolumn{1}{l}{67.0} &
  \multicolumn{1}{l}{18.0} &
  \multicolumn{1}{l}{24.0} &
  \multicolumn{1}{l}{31.9} &
  \multicolumn{1}{l|}{43.5} &
  56.0 &
  \multicolumn{1}{l}{46.0} &
  \multicolumn{1}{l}{48.0} &
  74.0 \\ \hline
\multirow{7}{*}{\rot{\textbf{Language models}}} &
  BERT-base &
  69.7 &
  42.5 &
  50.1 &
  45.5 &
  63.9 &
  62.4 &
  60.0 &
  58.8 &
  81 \\
 &
  BERT-large &
  \multicolumn{1}{l}{64.4} &
  \multicolumn{1}{l}{\textbf{51.5}} &
  \multicolumn{1}{l}{56.4} &
  \multicolumn{1}{l}{44.6} &
  \multicolumn{1}{l|}{65.1} &
  62.4 &
  \multicolumn{1}{l}{61.7} &
  \multicolumn{1}{l}{61.7} &
  84.3 \\
 &
  RB-base &
  68.5 &
  49.2 &
  55.8 &
  46.5 &
  66.2 &
  64.8 &
  66.7 &
  65.6 &
  85.9 \\
 &
  RB-large &
  \textbf{72.2} &
  48.9 &
  56.3 &
  \textbf{47.9} &
  \textbf{67.7} &
  66.1 &
  56.2 &
  58.3 &
  84.5 \\
 &
  BERTweet &
  66.9 &
  46.1 &
  52.7 &
  47.1 &
  66.9 &
  64.9 &
  65.6 &
  63.8 &
  85.2 \\
 &
  TimeLM-19 &
  71.1 &
  50.4 &
  \textbf{57.2} &
  47.7 &
  67.5 &
  \textbf{76.5} &
  \textbf{68.9} &
  \textbf{70.0} &
  \textbf{86.4} \\ \cline{2-11} 
 &
  \textit{TimeLM-21} &
  \textit{66.1} &
  \textit{54.2} &
  \textit{58.8} &
  \textit{47.1} &
  \textit{67.6} &
  \textit{73.9} &
  \textit{69.8} &
  \textit{70.1} &
  \textit{86.8}
\end{tabular}}
\caption{Macro average Precision (Pr), Recall (Rec), F1, and accuracy results in TweetTopic (temporal split). Jaccard Index (JI) is reported for the multi-label setting. }
\label{table:metrics}
\end{table}

\section{Analysis}
\label{analysis}

In this section, we analyse two important aspects of the TweetTopic dataset, mainly its temporal dimension (Section \ref{sec:temporal}) and the errors made by the systems (Section \ref{erroranalysis}). 

\subsection{Temporal analysis}
\label{sec:temporal}


The strong performance of TimeLM-21 provided evidence regarding the importance of an up-to-date training corpus. 
We continue our investigation by training the same set of models on a random split of the data (i.e., both training and test sets with tweets from 2019 to 2021). To make the results comparable, we created training and test sizes of the same size as the original temporal split.\footnote{While the distribution of labels may naturally be altered, this change is minimal, as we can recall from Figures \ref{fig:class_dist_multilabel} and \ref{fig:class_dist_singleclass}.}  


\begin{table*}[t]
\adjustbox{max width=\textwidth}{%
\begin{tabular}{|l|cc|cc|cc|cc|cc|cc|cc|cc|}
\hline
\multicolumn{1}{|c|}{\multirow{2}{*}{\textbf{Class}}} &
  \multicolumn{2}{c|}{\textbf{Random}} &
  \multicolumn{2}{c|}{\textbf{SVM}} &
  \multicolumn{2}{c|}{\textbf{BERT}} &
  \multicolumn{2}{c|}{\textbf{RB}} &
  \multicolumn{2}{c|}{\textbf{RB-large}} &
  \multicolumn{2}{c|}{\textbf{BERTweet}} &
  \multicolumn{2}{c|}{\textbf{TimeLM-19}} &
  \multicolumn{2}{c|}{\textbf{TimeLM-21}} \\ \cline{2-17} 
\multicolumn{1}{|c|}{} &
  \multicolumn{1}{c|}{\textbf{temp}} &
  \textbf{rand} &
  \multicolumn{1}{c|}{\textbf{temp}} &
  \textbf{rand} &
  \multicolumn{1}{c|}{\textbf{temp}} &
  \textbf{rand} &
  \multicolumn{1}{c|}{\textbf{temp}} &
  \textbf{rand} &
  \multicolumn{1}{c|}{\textbf{temp}} &
  \textbf{rand} &
  \multicolumn{1}{c|}{\textbf{temp}} &
  \textbf{rand} &
  \multicolumn{1}{c|}{\textbf{temp}} &
  \textbf{rand} &
  \multicolumn{1}{c|}{\textbf{temp}} &
  \textbf{rand} \\ \hline
arts \& culture &
  \multicolumn{1}{c|}{\textbf{8.6}} &
  8.3 &
  \multicolumn{1}{c|}{3.6} &
  \textbf{27.7} &
  \multicolumn{1}{c|}{17.8} &
  \textbf{35.9} &
  \multicolumn{1}{c|}{20.9} &
  \textbf{41.2} &
  \multicolumn{1}{c|}{28.0} &
  \textbf{44.0} &
  \multicolumn{1}{c|}{9.8} &
  \textbf{28.2} &
  \multicolumn{1}{c|}{21.3} &
  \textbf{39.1} &
  \multicolumn{1}{c|}{35.4} &
  \textbf{44.8} \\ \hline
business \& entrepreneurs &
  \multicolumn{1}{c|}{\textbf{8.7}} &
  7.4 &
  \multicolumn{1}{c|}{18.0} &
  \textbf{28.7} &
  \multicolumn{1}{c|}{\textbf{53.3}} &
  49.2 &
  \multicolumn{1}{c|}{56.7} &
  \textbf{57.1} &
  \multicolumn{1}{c|}{50.9} &
  56.5 &
  \multicolumn{1}{c|}{\textbf{59.7}} &
  54.5 &
  \multicolumn{1}{c|}{\textbf{58.6}} &
  55.3 &
  \multicolumn{1}{c|}{\textbf{56.3}} &
  54.0 \\ \hline
celebrity \& pop culture &
  \multicolumn{1}{c|}{22.8} &
  22.9 &
  \multicolumn{1}{c|}{22.3} &
  \textbf{41.7} &
  \multicolumn{1}{c|}{34.4} &
  \textbf{54.3} &
  \multicolumn{1}{c|}{47.2} &
  \textbf{52.7} &
  \multicolumn{1}{c|}{50.5} &
  \textbf{59.6} &
  \multicolumn{1}{c|}{43.7} &
  \textbf{54.9} &
  \multicolumn{1}{c|}{\textbf{48.6}} &
  47.8 &
  \multicolumn{1}{c|}{46.4} &
  \textbf{57.6} \\ \hline
diaries \& daily life &
  \multicolumn{1}{c|}{18.2} &
  \textbf{21.2} &
  \multicolumn{1}{c|}{25.8} &
  \textbf{34.4} &
  \multicolumn{1}{c|}{\textbf{45.2}} &
  44.0 &
  \multicolumn{1}{c|}{46.2} &
  \textbf{50.3} &
  \multicolumn{1}{c|}{43.5} &
  \textbf{49.3} &
  \multicolumn{1}{c|}{44.6} &
  \textbf{49.9} &
  \multicolumn{1}{c|}{44.5} &
  \textbf{51.2} &
  \multicolumn{1}{c|}{44.7} &
  \textbf{49.8} \\ \hline
family &
  \multicolumn{1}{c|}{3.5} &
  \textbf{6.3} &
  \multicolumn{1}{c|}{33.9} &
  \textbf{46.4} &
  \multicolumn{1}{c|}{47.2} &
  \textbf{48.3} &
  \multicolumn{1}{c|}{50.6} &
  \textbf{56.8} &
  \multicolumn{1}{c|}{52.8} &
  \textbf{63.4} &
  \multicolumn{1}{c|}{46.1} &
  \textbf{49.1} &
  \multicolumn{1}{c|}{46.4} &
  \textbf{55.2} &
  \multicolumn{1}{c|}{53.1} &
  \textbf{56.2} \\ \hline
fashion \& style &
  \multicolumn{1}{c|}{\textbf{4.8}} &
  4.1 &
  \multicolumn{1}{c|}{38.4} &
  \textbf{57.6} &
  \multicolumn{1}{c|}{52.8} &
  \textbf{74.8} &
  \multicolumn{1}{c|}{66.4} &
  \textbf{74.1} &
  \multicolumn{1}{c|}{66.4} &
  \textbf{77.4} &
  \multicolumn{1}{c|}{56.0} &
  \textbf{68.8} &
  \multicolumn{1}{c|}{\textbf{66.4}} &
  75.2 &
  \multicolumn{1}{c|}{67.2} &
  \textbf{75.2} \\ \hline
film tv \& video &
  \multicolumn{1}{c|}{\textbf{22.8}} &
  22.0 &
  \multicolumn{1}{c|}{47.3} &
  \textbf{58.6} &
  \multicolumn{1}{c|}{62.8} &
  \textbf{68.2} &
  \multicolumn{1}{c|}{64.4} &
  \textbf{71.4} &
  \multicolumn{1}{c|}{64.7} &
  \textbf{71.3} &
  \multicolumn{1}{c|}{66.8} &
  \textbf{69.2} &
  \multicolumn{1}{c|}{66.1} &
  \textbf{72.2} &
  \multicolumn{1}{c|}{65.4} &
  \textbf{70.6} \\ \hline
fitness \& health &
  \multicolumn{1}{c|}{6.6} &
  \textbf{9.3} &
  \multicolumn{1}{c|}{35.7} &
  \textbf{36.0} &
  \multicolumn{1}{c|}{\textbf{53.6}} &
  52.2 &
  \multicolumn{1}{c|}{52.4} &
  \textbf{53.2} &
  \multicolumn{1}{c|}{62.4} &
  \textbf{65.4} &
  \multicolumn{1}{c|}{\textbf{48.2}} &
  38.7 &
  \multicolumn{1}{c|}{\textbf{55.7}} &
  42.2 &
  \multicolumn{1}{c|}{58.6} &
  52.6 \\ \hline
food \& dining &
  \multicolumn{1}{c|}{3.5} &
  \textbf{4.6} &
  \multicolumn{1}{c|}{25.0} &
  \textbf{41.7} &
  \multicolumn{1}{c|}{\textbf{70.1}} &
  68.2 &
  \multicolumn{1}{c|}{\textbf{75.1}} &
  75.3 &
  \multicolumn{1}{c|}{\textbf{79.3}} &
  68.2 &
  \multicolumn{1}{c|}{\textbf{74.5}} &
  65.7 &
  \multicolumn{1}{c|}{\textbf{75.4}} &
  70.7 &
  \multicolumn{1}{c|}{\textbf{80.4}} &
  71.6 \\ \hline
gaming &
  \multicolumn{1}{c|}{6.9} &
  \textbf{7.5} &
  \multicolumn{1}{c|}{31.8} &
  \textbf{45.0} &
  \multicolumn{1}{c|}{57.4} &
  \textbf{61.2} &
  \multicolumn{1}{c|}{58.4} &
  \textbf{61.4} &
  \multicolumn{1}{c|}{63.8} &
  \textbf{69.1} &
  \multicolumn{1}{c|}{66.1} &
  \textbf{67.6} &
  \multicolumn{1}{c|}{64.6} &
  \textbf{69.2} &
  \multicolumn{1}{c|}{64.8} &
  71.2 \\ \hline
learning \& educational &
  \multicolumn{1}{c|}{4.2} &
  \textbf{4.5} &
  \multicolumn{1}{c|}{13.0} &
  \textbf{13.9} &
  \multicolumn{1}{c|}{38.2} &
  \textbf{43.2} &
  \multicolumn{1}{c|}{\textbf{49.5}} &
  48.7 &
  \multicolumn{1}{c|}{49.8} &
  45.8 &
  \multicolumn{1}{c|}{\textbf{42.9}} &
  36.2 &
  \multicolumn{1}{c|}{\textbf{49.3}} &
  47.1 &
  \multicolumn{1}{c|}{\textbf{48.9}} &
  47.0 \\ \hline
music &
  \multicolumn{1}{c|}{24.7} &
  \textbf{25.5} &
  \multicolumn{1}{c|}{76.1} &
  \textbf{81.8} &
  \multicolumn{1}{c|}{83.6} &
  \textbf{86.0} &
  \multicolumn{1}{c|}{86.0} &
  \textbf{87.1} &
  \multicolumn{1}{c|}{87.4} &
  \textbf{88.1} &
  \multicolumn{1}{c|}{86.9} &
  \textbf{87.2} &
  \multicolumn{1}{c|}{\textbf{88.1}} &
  87.8 &
  \multicolumn{1}{c|}{86.9} &
  \textbf{88.2} \\ \hline
news \& social concern &
  \multicolumn{1}{c|}{39.3} &
  \textbf{39.9} &
  \multicolumn{1}{c|}{69.8} &
  \textbf{76.9} &
  \multicolumn{1}{c|}{\textbf{83.8}} &
  \textbf{83.8} &
  \multicolumn{1}{c|}{83.9} &
  \textbf{84.6} &
  \multicolumn{1}{c|}{85.5} &
  \textbf{85.9} &
  \multicolumn{1}{c|}{83.5} &
  \textbf{84.3} &
  \multicolumn{1}{c|}{84.4} &
  \textbf{86.2} &
  \multicolumn{1}{c|}{84.5} &
  \textbf{85.0} \\ \hline
other hobbies &
  \multicolumn{1}{c|}{\textbf{10.5}} &
  9.6 &
  \multicolumn{1}{c|}{4.2} &
  \textbf{15.0} &
  \multicolumn{1}{c|}{\textbf{27.0}} &
  23.6 &
  \multicolumn{1}{c|}{25.0} &
  \textbf{28.4} &
  \multicolumn{1}{c|}{31.7} &
  \textbf{35.4} &
  \multicolumn{1}{c|}{\textbf{23.1}} &
  21.5 &
  \multicolumn{1}{c|}{27.7} &
  \textbf{30.3} &
  \multicolumn{1}{c|}{31.1} &
  26.2 \\ \hline
relationships &
  \multicolumn{1}{c|}{6.4} &
  \textbf{7.3} &
  \multicolumn{1}{c|}{13.7} &
  \textbf{36.3} &
  \multicolumn{1}{c|}{30.8} &
  \textbf{35.2} &
  \multicolumn{1}{c|}{37.6} &
  \textbf{51.8} &
  \multicolumn{1}{c|}{39.3} &
  \textbf{56.8} &
  \multicolumn{1}{c|}{36.8} &
  \textbf{51.2} &
  \multicolumn{1}{c|}{35.3} &
  \textbf{51.6} &
  \multicolumn{1}{c|}{44.5} &
  \textbf{54.0} \\ \hline
samples avg &
  \multicolumn{1}{c|}{13.8} &
  \textbf{14.3} &
  \multicolumn{1}{c|}{57.0} &
  \textbf{63.7} &
  \multicolumn{1}{c|}{\textbf{70.3}} &
  72.0 &
  \multicolumn{1}{c|}{73.1} &
  \textbf{74.2} &
  \multicolumn{1}{c|}{74.4} &
  \textbf{76.4} &
  \multicolumn{1}{c|}{\textbf{73.8}} &
  73.2 &
  \multicolumn{1}{c|}{74.3} &
  \textbf{75.2} &
  \multicolumn{1}{c|}{74.7} &
  \textbf{75.2} \\ \hline
science \& technology &
  \multicolumn{1}{c|}{8.3} &
  \textbf{9.3} &
  \multicolumn{1}{c|}{17.4} &
  \textbf{35.8} &
  \multicolumn{1}{c|}{45.9} &
  \textbf{50.3} &
  \multicolumn{1}{c|}{54.2} &
  \textbf{56.4} &
  \multicolumn{1}{c|}{52.1} &
  \textbf{59.4} &
  \multicolumn{1}{c|}{46.9} &
  \textbf{53.2} &
  \multicolumn{1}{c|}{50.5} &
  \textbf{56.0} &
  \multicolumn{1}{c|}{50.2} &
  \textbf{52.1} \\ \hline
sports &
  \multicolumn{1}{c|}{\textbf{36.6}} &
  34.8 &
  \multicolumn{1}{c|}{82.2} &
  \textbf{89.1} &
  \multicolumn{1}{c|}{93.1} &
  \textbf{93.2} &
  \multicolumn{1}{c|}{\textbf{94.8}} &
  94.2 &
  \multicolumn{1}{c|}{94.6} &
  \textbf{95.4} &
  \multicolumn{1}{c|}{\textbf{95.4}} &
  94.4 &
  \multicolumn{1}{c|}{\textbf{95.6}} &
  94.8 &
  \multicolumn{1}{c|}{\textbf{95.2}} &
  94.8 \\ \hline
travel \& adventure &
  \multicolumn{1}{c|}{2.2} &
  \textbf{3.5} &
  \multicolumn{1}{c|}{\textbf{17.7}} &
  9.8 &
  \multicolumn{1}{c|}{\textbf{21.7}} &
  20.6 &
  \multicolumn{1}{c|}{41.5} &
  \textbf{47.7} &
  \multicolumn{1}{c|}{46.3} &
  \textbf{59.9} &
  \multicolumn{1}{c|}{\textbf{38.5}} &
  0.0 &
  \multicolumn{1}{c|}{\textbf{57.1}} &
  56.0 &
  \multicolumn{1}{c|}{52.2} &
  54.7 \\ \hline
youth \& student life &
  \multicolumn{1}{c|}{1.7} &
  \textbf{2.9} &
  \multicolumn{1}{c|}{2.9} &
  \textbf{12.4} &
  \multicolumn{1}{c|}{33.3} &
  \textbf{44.6} &
  \multicolumn{1}{c|}{49.2} &
  \textbf{52.4} &
  \multicolumn{1}{c|}{21.0} &
  \textbf{46.0} &
  \multicolumn{1}{c|}{31.6} &
  \textbf{35.2} &
  \multicolumn{1}{c|}{\textbf{50.4}} &
  43.6 &
  \multicolumn{1}{c|}{50.8} &
  \textbf{51.0} \\ \hline
\textbf{macro avg} &
  \multicolumn{1}{c|}{12.6} &
  \textbf{13.2} &
  \multicolumn{1}{c|}{30.5} &
  \textbf{41.5} &
  \multicolumn{1}{c|}{50.1} &
  \textbf{54.6} &
  \multicolumn{1}{c|}{55.8} &
  \textbf{60.3} &
  \multicolumn{1}{c|}{56.3} &
  \textbf{63.0} &
  \multicolumn{1}{c|}{52.7} &
  \textbf{53.1} &
  \multicolumn{1}{c|}{57.2} &
  \textbf{59.6} &
  \multicolumn{1}{c|}{58.8} &
  \textbf{60.9} \\ \hline
\end{tabular}}%
\caption{Macro average F1 scores for the multi-label setting when using temporal (temp) and random (rand) split. Highlighted with bold is the best score for each model.}
\label{tab:f1_temporal-random}
\end{table*}

\begin{figure}[t]
    \centering
    \includegraphics[width=7.5cm]{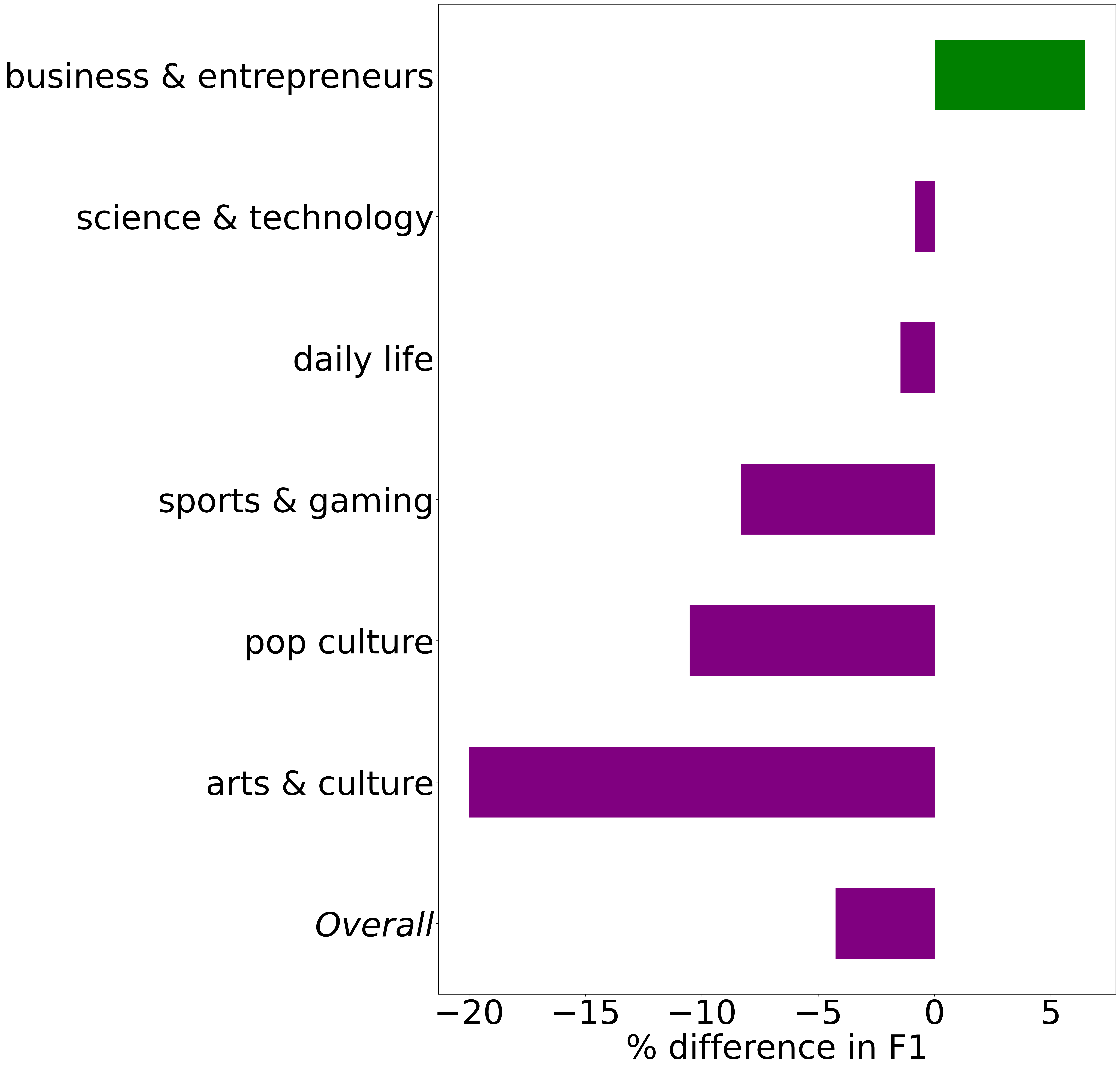}
    \caption{Relative (\%) differences in F1 scores when \textit{TimeLM-19} is trained in a temporal and in a random setting for the single-label setting. Negative values indicate that when using the temporal split the model's performance decreases. }
    \label{fig:performance_diff_single}
\end{figure}

Table \ref{tab:f1_temporal-random} displays the F1 scores, while using a multi-label setting for each class in both the temporal and random splits. Every model tested performs better when trained using information from both time periods, i.e using random split. Taking into account that in both splits the distribution of classes is similar (Figure \ref{fig:class_dist_multilabel}), we can assume that the temporal differences in the data provide useful information. It is worth noting that the "specialized" Twitter models display a more robust performance regarding the training data used. In particular, there are  8, 9 and 4 topics where BERTweet, TimeLM-19, and TimeLM-21 respectively perform better while using the temporal split in contrast to 3 and 1 of RoBERTa base and large respectively (models that have a similar architecture).

We continue our analysis by investigating in more detail TimeLM-19's results, which is the best performing model according to the evaluation (Section \ref{evaluation}).  
Figure \ref{fig:performance_diff_single} displays the TimeLM-19 performance differences between the temporal and random splits on the single-label setting. In general, results are overall better in the random split, with an overall relative decrease of 4.3\% in Macro-F1 for the temporal split. 
The largest decrease in performance is observed for the \textit{arts \& culture} topic in both settings, which can be attributed to a 
fast evolving vocabulary. In contrast, \textit{business \& entrepreneurs} does not see any decreased in performance in both settings, and results are even slightly better on the temporal split.\footnote{In the Appendix we provide a detailed analysis by quarter, in order to better understand the temporal aspect. The results confirm how the performance of \textit{arts \& culture} decreases over time, while for the rest of the topics the trend is unclear.}

\begin{figure}[t]
    \centering
    \includegraphics[width=7.5cm]{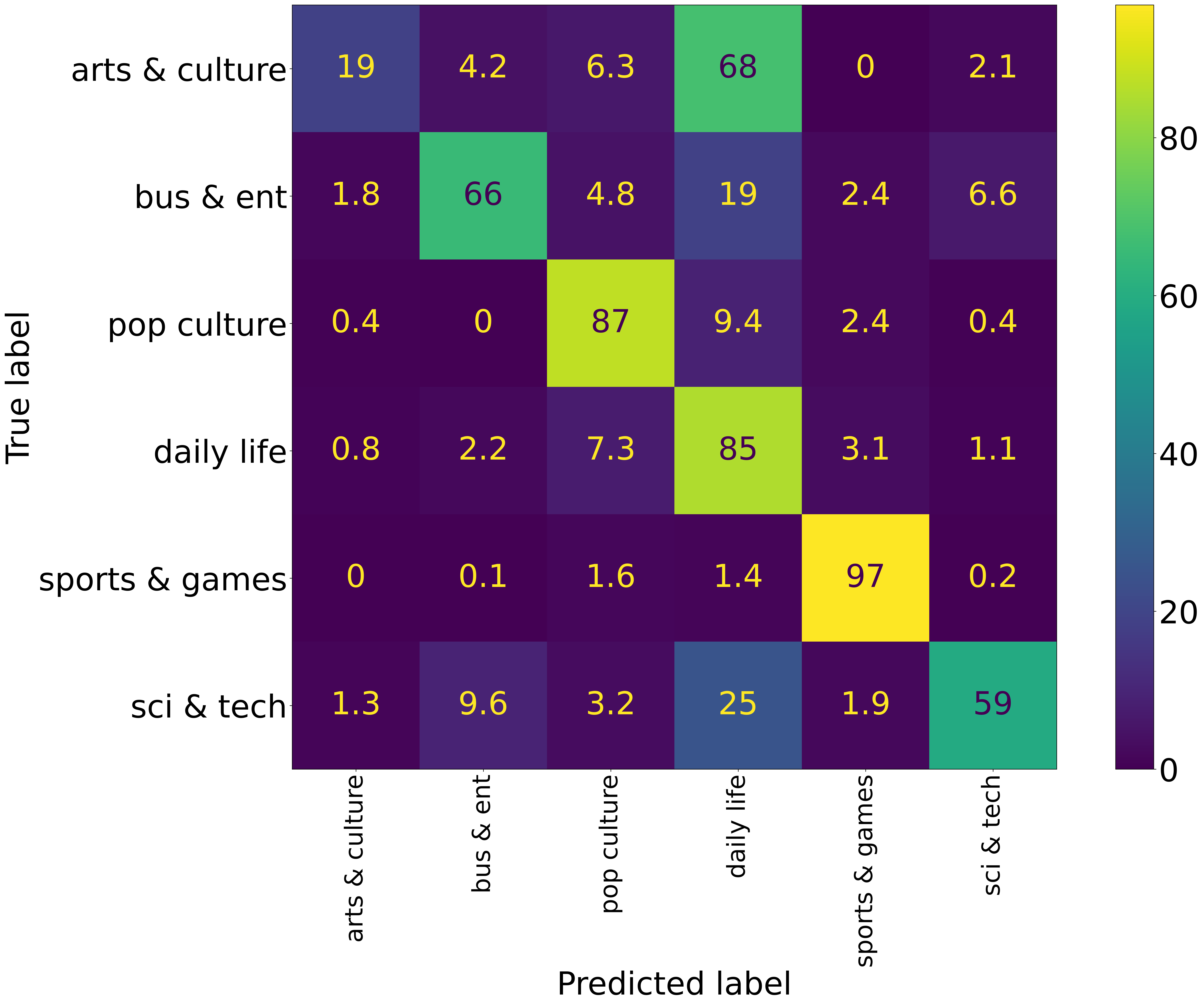}
    \caption{Confusion matrix of the TimeLM-19 results for the single-label setting. The values displayed are normalized  by row.}
    \label{fig:confusion_matrix}
\end{figure}

\subsection{Error analysis}
\label{erroranalysis}
To better understand the nature of errors made by language models, Figure \ref{fig:confusion_matrix} shows a confusion matrix for the best-performing TimeLM-19 model in the single-label setting. 
The model seems to struggle with tweets assigned to the \textit{arts \& culture} topic with 68\% of them being misclassified as \textit{daily life}.
These errors include entries such as ``Happy Day of the Dead 2020! \#GoogleDoodle'' 
or ``Gifts of love are the ingredients of a \#MerryChristmas  Give 
your loved ones a physical/virtual crypto gift card within the \{\{USERNAME\}\} app''.  
While these tweets revolve around religious/cultural holidays, one might also associate them to daily life events, which also shows the challenging nature of this dataset. 
Another topic that is frequent misclassified is \textit{science \& technology}, with 41\% of the tweets being assigned to the wrong topic. When looking at the errors we identify tweets such as ``Bill Gates-Funded Company Releases Genetically Modified Mosquitoes in US'', classified as \textit{business \& entrepreneurs}, and ``Monday’s Google Doodle Celebrates Jupiter And Saturn On The Winter Solstice via Forbes'', classified as \textit{daily life}. 
In other cases, further investigation would be required to understand the source of the mistakes, e.g., ``A year ago we looked at PE10s across the world on {{URL}} The latest Weekly Macro Themes takes a look at how the Euro Area stacks up now.'' was classified as \textit{sports} instead \textit{business \& entrepreneurs}. The nature of these types of error, as well as the relatively low performance of models compares to other topic classification datasets, suggest that there is ample room for improvement.

When considering the multi-label setting, there are topics with high percentage of errors such as \textit{celebrity \& pop culture} and \textit{diaries \& daily life}. There are entries like ``Anyone else notice \{\@O Shea Jack Nichol son\@\} hasn’t tweeted about the Lakers making the conference finals? Weird. You good man?'' where the model correctly classifies it as \textit{sports} but fails to classify it as \textit{celebrity \& pop culture}, being probably unaware of the celebrity status of the person being mentioned. The \textit{diaries \& daily life} topic seems to be particular confusing for the model and fails to identify it in tweets such as ``Lost all my bets on the Kentucky Derby today but scored a tee time at \{\{USERNAME\}\} Black course next weekend I’d say I came out a winner.'', and ``Faceing difficulty while login to internet banking for the 1st time using Id and password provided in the welcome kit didn t expected this from such a good bank \{\@Canara Bank\@\}'',
even though they are correctly assigned the \textit{sports} and \textit{business \& entrepreneurs} topics, respectively. 


\section{Conclusions \& Future Work}


In this paper we presented TweetTopic, the first large-scale dataset for tweet topic classification. Given the prominence of social media in recent times, this dataset can help build supervised models for clustering and organising the online content. 
The curated set of topics contains a diverse and broad set of categories that cover most topics present in social platform data. This dataset can further motivate research on the evolution of these initial topics on social platforms, i.e., the extension of the existing categorization to new topics or subtopics that will emerge and fade over time due to user engagement. 
Moreover, TweetTopic has been shown to be relatively resilient to temporal changes, and it offers easily interpretable results. Based on these contributions, we believe that this dataset will be useful for a significant number of researchers and practitioners working on social media, including Computational Social Science and Data Mining experts, given the relevance of the topic for extracting information and understanding online behavior.

Finally, while this first iteration of TweetTopic focuses on English, our aim is to apply the same methodology to other languages, for which our guidelines and process to construct the dataset described in Section \ref{data} can serve as the main basis.

\section*{Acknowledgements}

Jose Camacho-Collados is supported by a UKRI Future Leaders Fellowship.



\bibliography{anthology,custom}
\bibliographystyle{acl_natbib}

\appendix

\section{Tweet filtering}
\begin{figure}[hbt]
    \centering
    \includegraphics[width=7.5cm]{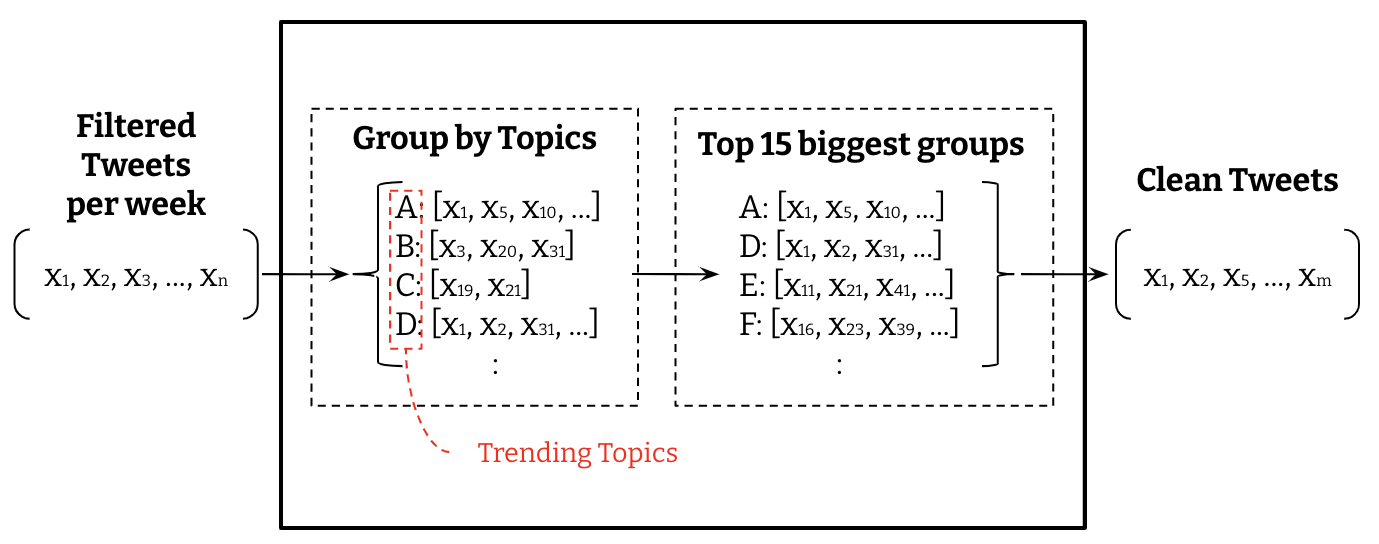}
    \caption{Weekly trend filtering to remove tweets that are irrelevant to the popular topics in each week.}
    \label{fig:trending}
\end{figure}

\begin{figure}[hbt]
    \centering
    \includegraphics[width=7.5cm]{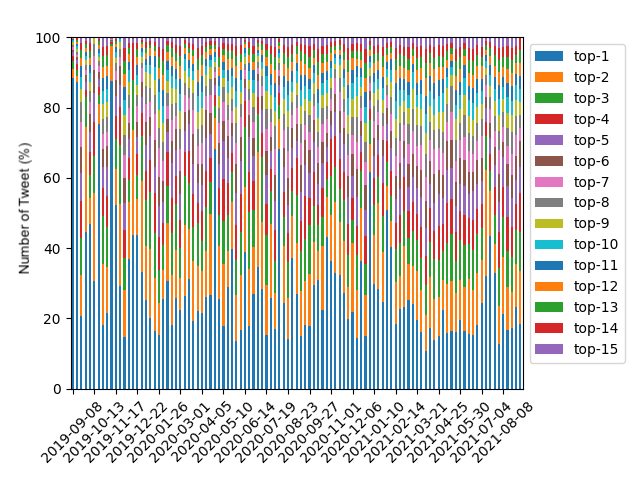}
    \caption{Ratio (\%) of tweets in each of top 15 trending keywords for every week.}
    \label{fig:trend_distribution}
\end{figure}

Figure \ref{fig:trending} illustrates the weekly trend filtering pipeline utilized.
Figure \ref{fig:trend_distribution} displays the weekly distribution of the top 15 trending topics used to query the raw tweets. 

\section{Annotation Interface}
\label{sec:annotationinterface}

\begin{figure*}[t]
    \centering
    \includegraphics[width=15cm]{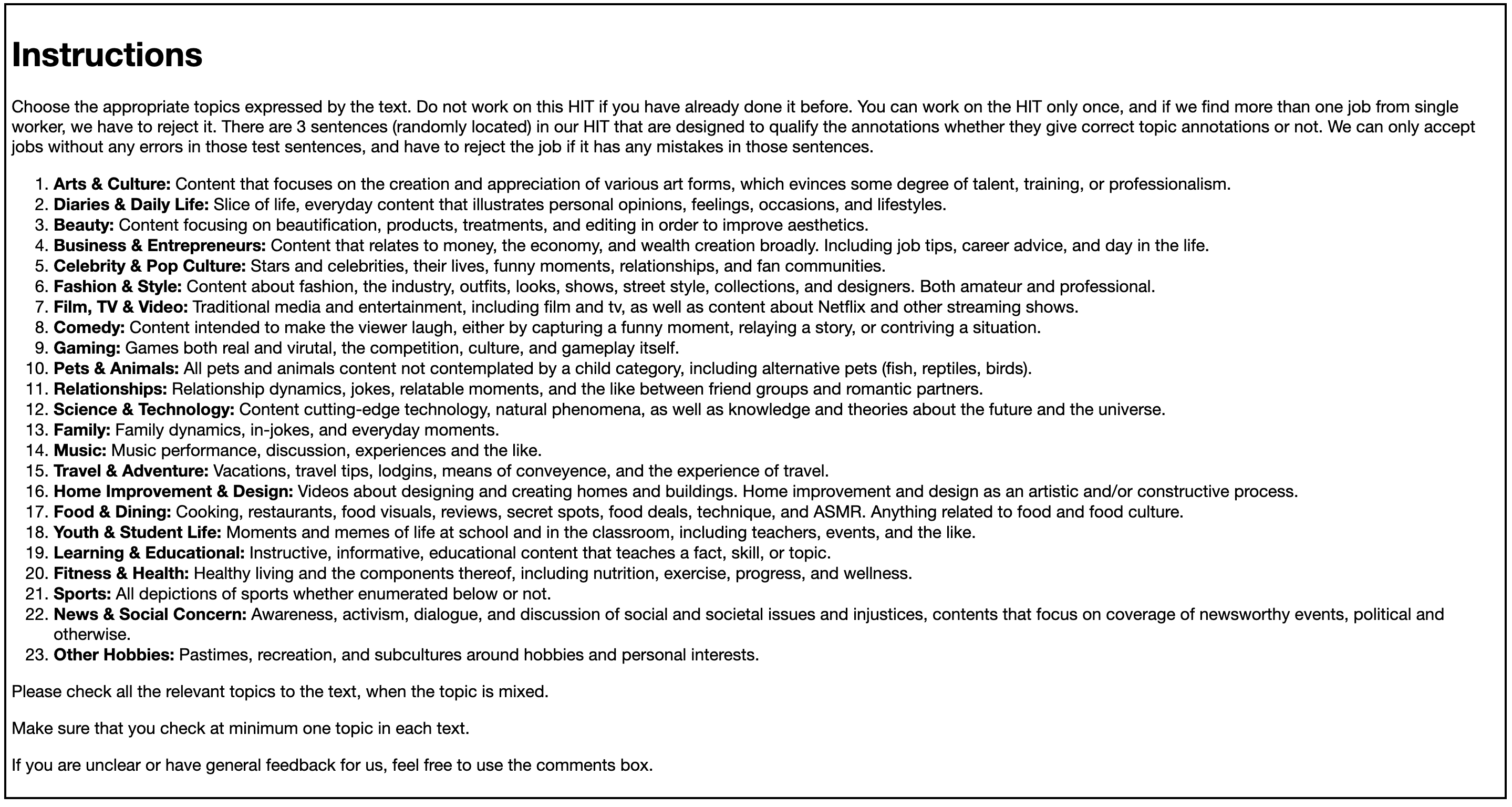}
    \caption{The instructions shown to the annotators during the annotation phase.}
    \label{fig:instruction}
\end{figure*}

\begin{figure}[t]
    \centering
    \includegraphics[width=7.5cm]{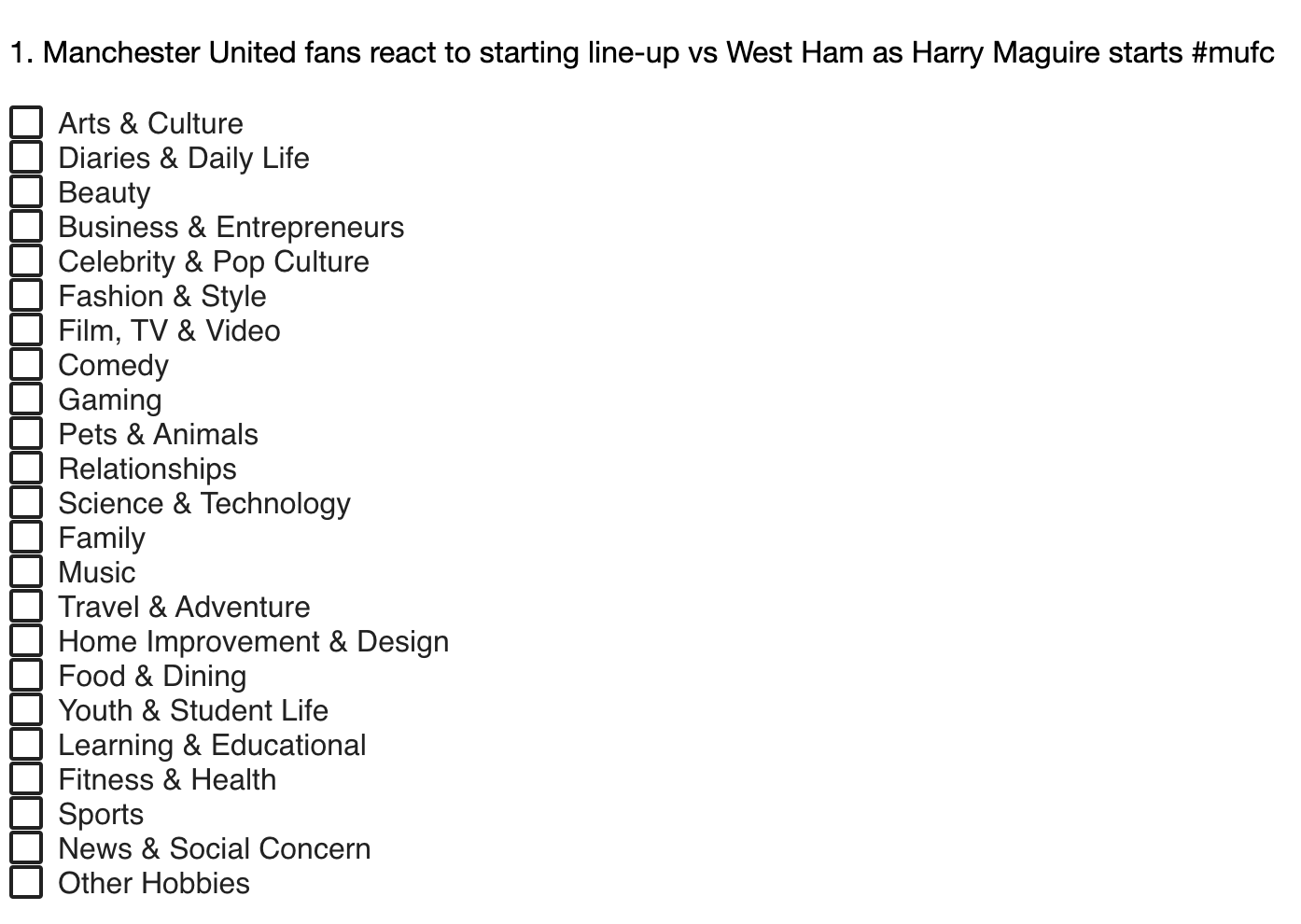}
    \caption{Tweet classification annotation interface. Annotators are allowed to select multiple topics.}
    \label{fig:interface}
\end{figure}

Figure \ref{fig:interface} presents our annotation interface. Figure \ref{fig:instruction} displays the instructions provided to annotators along with a small description of each topic.

\section{Evaluation Results}

\paragraph{Hyperparameters.} Language models are trained using a batch size of 8 for 20 epochs, while utilizing an Adam optimizer \cite{loshchilov2017decoupled} with learning rate $2e^-5$ and a weight decay of $0.01$. Furthermore, an early stop callback  terminates the training process after 3 epochs without performance improvement. Finally, for the single-label experiments cross entropy loss along with a softmax activation function were used, while for the multi-label setting binary cross entropy loss and a sigmoid activation for each of the 19 topics are used.

\begin{figure}[hbt!]
    \centering
    \includegraphics[width=7.5cm]{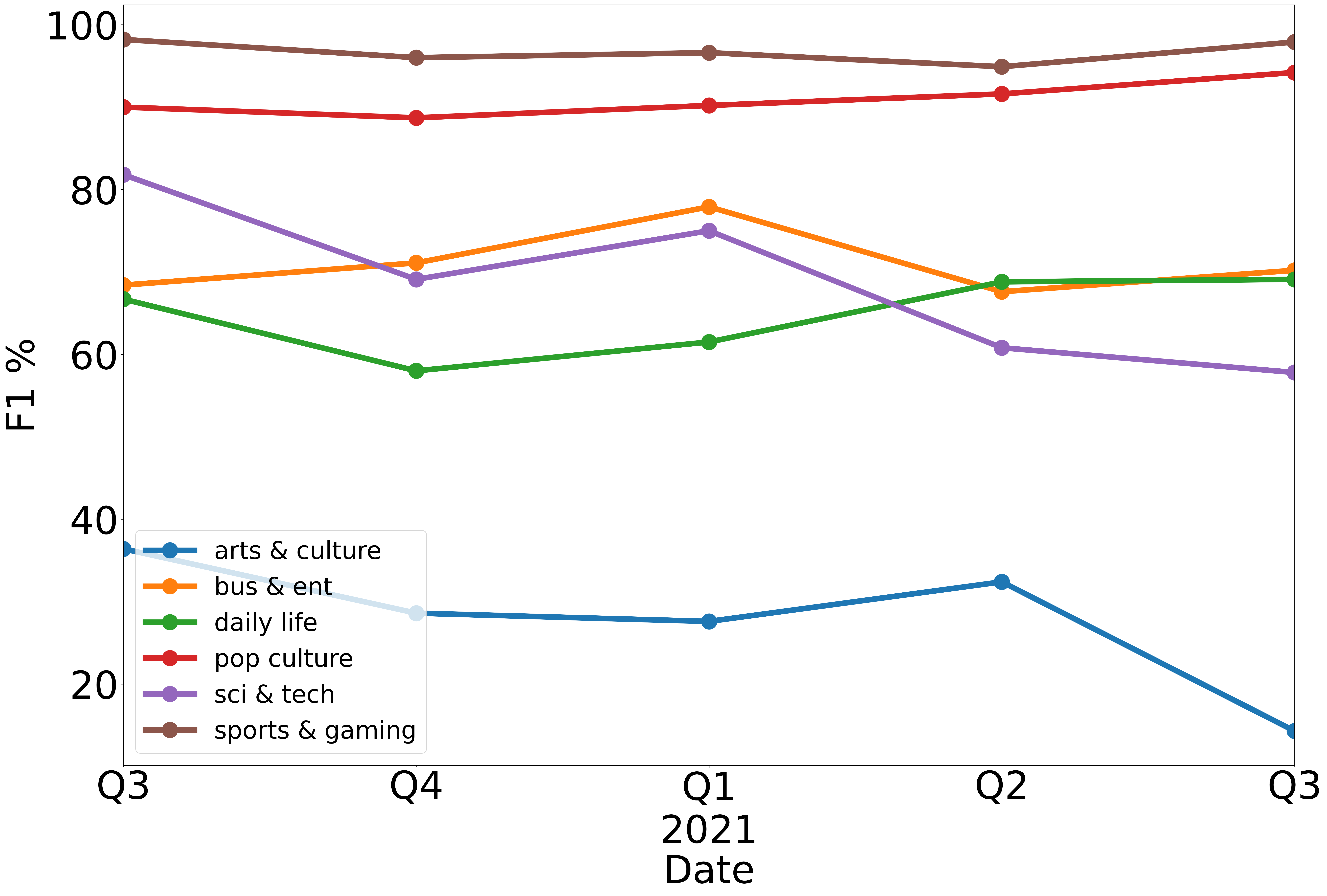}
    \caption{F1 performance of TimeLM-19 through time (single-label setting).}
    \label{fig:f1_over_time_single}
\end{figure}
\begin{figure}[t]
    \centering
    \includegraphics[width=7.5cm]{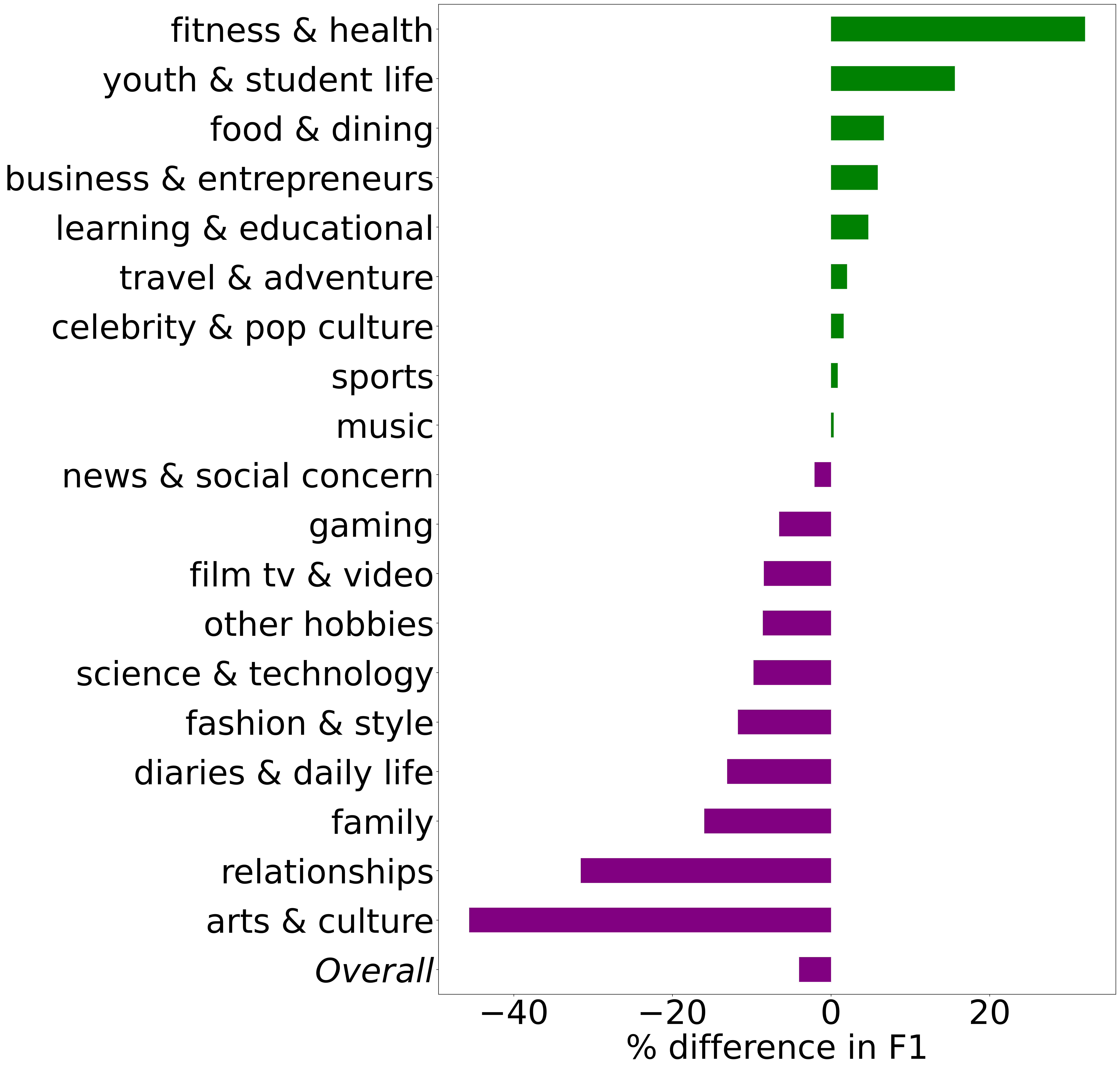}
    \caption{Relative (\%) differences in F1 scores when \textit{TimeLM-19} is trained in a temporal and in a random setting for the multi-label setting. Negative values indicate that when using the temporal split the model's performance decreases. }
    \label{fig:performance_diff_multi}
\end{figure}

\paragraph{Analysis by quarter.} In order to get a better understanding of the evolution of the corpus and identify potential performance decays due to temporal differences we inspect the performance of TimeLM-19 in each quarter (i.e., three months) of the temporal's split test-set. Figure \ref{fig:f1_over_time_single} displays the F1 scores of each class (single-label setting) for each quarter of the time period tested. While most topics do not seem to be greatly affected by time, we can indeed observe a performance drop in \textit{arts \& culture}, which is the topic more affected by the temporal variable. Figure \ref{fig:performance_diff_multi} illustrates the relative differences in F1 scores for each class in the multi-label setting, when TimeLM-19 is trained using the temporal split and when trained on the random split.

\begin{figure*}[t]
    \centering
    \includegraphics[width=.9\textwidth]{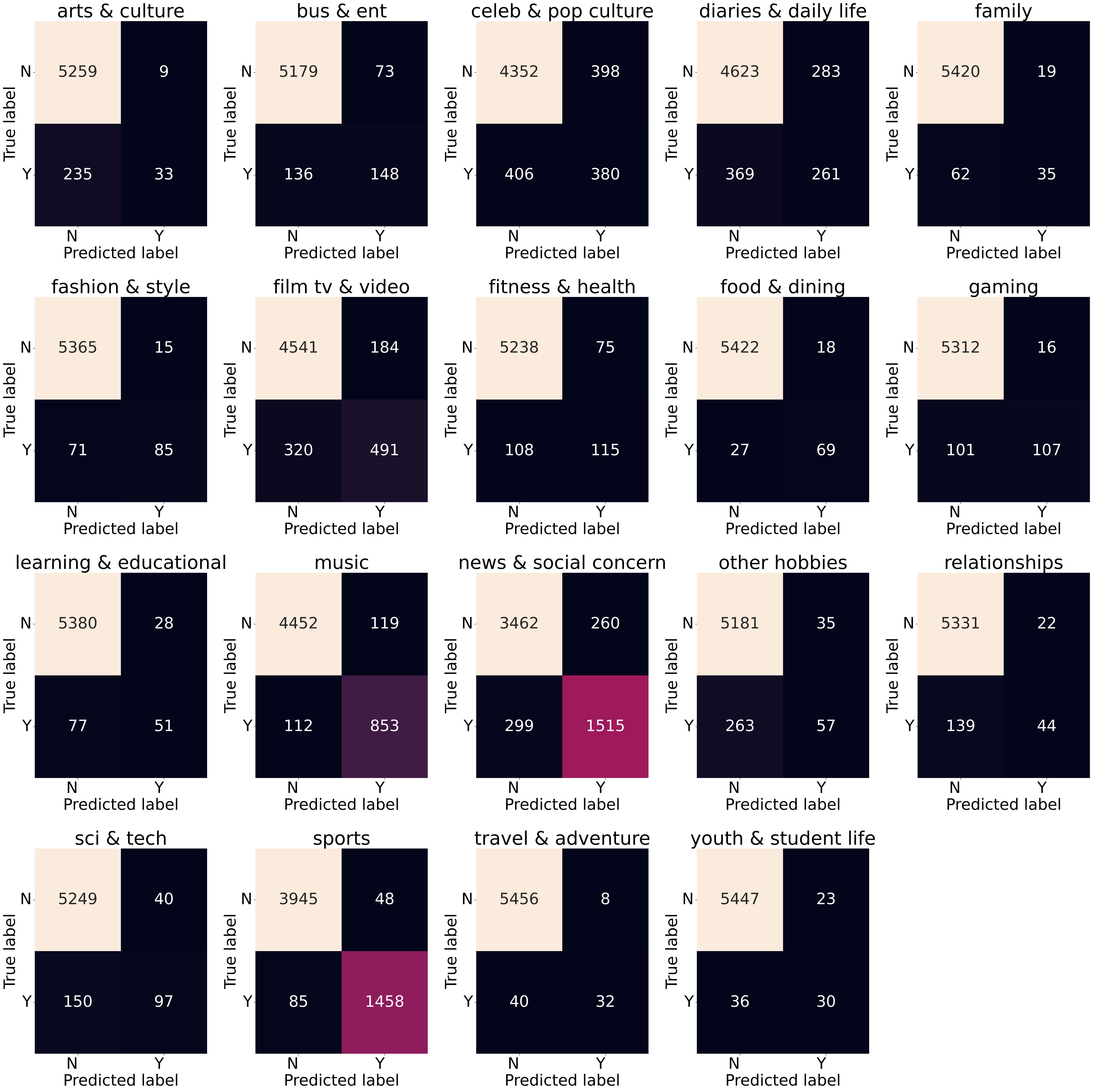}
    \caption{Confusion-matrix of TimeLM-19 (multi-label setting).}
    \label{fig:confusion_matrix_multi}
\end{figure*}

\paragraph{Confusion matrices.} Figure \ref{fig:confusion_matrix_multi} displays the confusion matrices for TimeLM-19 when trained in the multi-label setting using the temporal split.

\end{document}